\documentclass{bmvc2k}
\usepackage{amssymb}

\title{A Closer Look at Few-Shot Video Classification: A New Baseline and Benchmark}

\addauthor{Zhenxi Zhu}{zhuzhenxi@smail.nju.edu.cn}{1}
\addauthor{Limin Wang}{lmwang@nju.edu.cn}{1}
\addauthor{Sheng Guo}{guosheng.guosheng@alibaba-inc.com}{2}
\addauthor{Gangshan Wu}{gswu@nju.edu.cn}{1}

\addinstitution{
 State Key Laboratory for Novel Software Technology, Nanjing University, China
}
\addinstitution{
 MyBank, Ant Group, China
}

\runninghead{Zhu, Wang, Guo, Wu}{A Closer Look at Few-Shot Video Classification}

\def\etal{\emph{et al}\bmvaOneDot}

\begin{document}

\maketitle

\begin{abstract}

The existing few-shot video classification methods often employ a meta-learning paradigm by designing customized temporal alignment module for similarity calculation. While significant progress has been made, these methods fail to focus on learning effective representations, and heavily rely on the ImageNet pre-training, which might be unreasonable for the few-shot recognition setting due to semantics overlap. In this paper, we aim to present an in-depth study on few-shot video classification by making three contributions. First, we perform a consistent comparative study on the existing metric-based methods to figure out their limitations in representation learning. Accordingly, we propose a simple classifier-based baseline without any temporal alignment that surprisingly outperforms the state-of-the-art meta-learning based methods. Second, we discover that there is a high correlation between the novel action class and the ImageNet object class, which is problematic in the few-shot recognition setting. Our results show that the performance of training from scratch drops significantly, which implies that the existing benchmarks cannot provide enough base data. Finally, we present a new benchmark with more base data to facilitate future few-shot video classification without pre-training. The code will be made available at \url{https://github.com/MCG-NJU/FSL-Video}.

\end{abstract}

\section{Introduction}
\label{sec:intro}
Deep learning methods have achieved great success on the task of video action classification~\cite{C3D,TSN,I3D,slowfast}. Generally, a large amount of labeled data is required to successfully train such deep models for video classification. When generalizing to unseen classes, it still requires hundreds of labeled samples to retrain these models to recognize novel classes. This labeling requirement often prevents these deep models from being efficiently deployed in an open-world setting. Therefore, few-shot video recognition is becoming popular in realistic scenarios, which aims to recognize novel classes with limited labeled videos.

Few-shot learning has been widely studied in various research areas such as computer vision \cite{matching, CMN}, natural language processing \cite{nlp}, and bioimaging \cite{biology,zhou2020one}. One promising direction is the meta-learning paradigm \cite{closer} where transferable knowledge is learned from a collection of tasks (or episodes) to prevent over-fitting and improve generalization. Inspired by metric learning methods \cite{proto,relation}, the existing few-shot video classification methods \cite{TARN,OTAM,TRX,implicit,PAL,CMOT} usually compare the similarity of different videos in the feature space for classification. The essential difference between videos and images is the extra temporal dimension, which makes it insufficient to represent a whole video as a single feature vector. Therefore, many video-specific temporal alignment methods \cite{OTAM,CMOT,TRX,implicit} have been proposed to solve 
this problem. 

There are two major limitations in these meta-learning methods for few-shot video classification. {\bf First}, they often focus on designing effective temporal aggregation methods to combine local features~\cite{CMN,OTAM,TRX,TARN,CMOT} into video-level representations and optimizing the whole pipeline in a meta-learning way. However, they often overlook the importance of visual feature representation itself. We find that the simple pre-training and fine-tuning paradigm is even never explored for the task of few-shot video classification. {\bf Second}, the existing few-shot video classification methods all use pre-trained weights (e.g., pre-trained on ImageNet) to initialize the network parameters. However, this pre-training will violate the basic assumption of few-shot learning that the novel classes cannot be seen during meta-training. We find that ImageNet contains very high-related classes to those novel classes during meta-testing, which makes ImageNet pre-training unreasonable and problematic.

In this paper, we aim to present an in-depth study on few-shot video classification and hopefully provide new insights on this problem. First, we fairly compare several meta-learning based few-shot video classification methods with different temporal alignment, which provides a solid study on the importance of temporal alignment. We find that in this fair comparison, the performance gap between different temporal alignment methods is reduced,  in particular for 5-shot recognition. Then, we propose a simple classifier-based baseline method and surprisingly find that this simple baseline with weight imprinting and dropout achieves astoundingly good performance compared with those previous meta-learning methods. Finally, we analyze the effect of ImageNet pre-training on few-shot video classification and figure out that the current few-shot video recognition benchmarks cannot provide a reasonable number of base samples for training from scratch. To facilitate the future few-shot video classification without pre-training, we present a new benchmark based on the Kinetics dataset. The main contribution of this paper is summarized as follows:
\begin{itemize}
    \item We conduct consistent experiments to compare meta-learning based few-shot video classification methods and analyze their weakness in representation learning. Based on this analysis, we present a classifier-based baseline with weight imprinting and dropout, which significantly outperforms other classifier-based methods and previous state-of-the-art meta-learning methods on both the Kinetics and the Something-Something-V2 datasets.
    \item We specifically investigate the effect of ImageNet pre-training on few-shot video classification. This pre-training may violate the assumption of few-shot learning that novel classes cannot be seen before meta-testing stage. Our study on self-supervised pre-training shows that ImageNet contains highly related categories with unseen action classes, and without this pre-training, the performance of all methods drops significantly.
    \item Furthermore, we find that the existing few-shot video classification benchmark is limited in the number of training samples in the base set, and fail to provide a reasonable training set for learning from scratch. Therefore, we establish a new benchmark based on Kinetics, termed as {\em complete-Kinetics}, and hopefully to facilitate the future few-shot video classification without ImageNet pre-training.
\end{itemize}

\section{Related Work}

{\bf Few-Shot Learning.} 
Meta-learning paradigm, which trains the model with few-shot tasks constructed from training data, has been widely used in few-shot learning. Methods in this paradigm can be roughly divided into initialization based methods and metric based methods. Initialization based methods aimed to learn good model initialization so that the classifiers for novel classes can be learned quickly with one or several gradient updates \cite{finn2017, finn2018}. Metric based methods have become the most commonly used methods in few-shot video classification \cite{OTAM,TARN,CMN}, which addressed the few-shot classification problem by comparing samples from the query set and the support set. 
Another branch of works, which trained a model on the training set and fine-tuned with the few data samples of the novel classes, also achieved competitive performance even with a simple architecture. Chen \etal~\cite{closer} replaced the linear classifier with a distance-based classifier to reduce intra-class variation. Dhillon \etal~\cite{baseline} proposed a support-based initialization and transductive fine-tuning baseline, which preserved the linear classifier with support weight initialization. Chen \etal~\cite{metabaseline} removed the linear classifier only in the testing stage and measured cosine similarity between the query and the support samples.

\noindent {\bf Video Classification.}
Video classification has been extensively studied in the past few years. The methods can be divided into two categories: 2D convolution based methods \cite{TSN,tam,tdn} and 3D convolution based methods \cite{slowfast,C3D,I3D}. TSN \cite{TSN} and C3D \cite{C3D} became the most popular methods in few-shot video classification \cite{OTAM,CMN,TARN,CMOT}. TSN sparsely and uniformly sampled a fixed number of frames for a 2D backbone. C3D utilized 3D spatio-temporal convolutional filters to extract segment-level features. 

\noindent {\bf Temporal Alignment.} 
Temporal alignment is a method to match two video sequences in the temporal dimension. Since metric based methods need to compare samples in feature space, various alignment methods have been proposed to combine local features (e.g., features extracted by TSN or C3D) before comparison. Temporal alignment can be both explicit \cite{OTAM,CMOT} or implicit \cite{CMN,implicit}. The goal of temporal alignment is to obtain better global features for comparison. On the other hand, relation module \cite{relation} has been designed for better comparison of two different global features, which was commonly used in few-shot image classification. Since the boundary between temporal alignment and relation module is vague, we distinguish them by whether the similarity can be calculated directly. Specifically, temporal alignment ends at the video-level feature representations or aligned features of two different videos in the temporal dimension.

\section{Few-Shot Video Classification}

\subsection{Meta-learning Based Methods Revisited} \label{meta_methods}

\begin{figure}[t]

\centering
\includegraphics[scale=0.44]{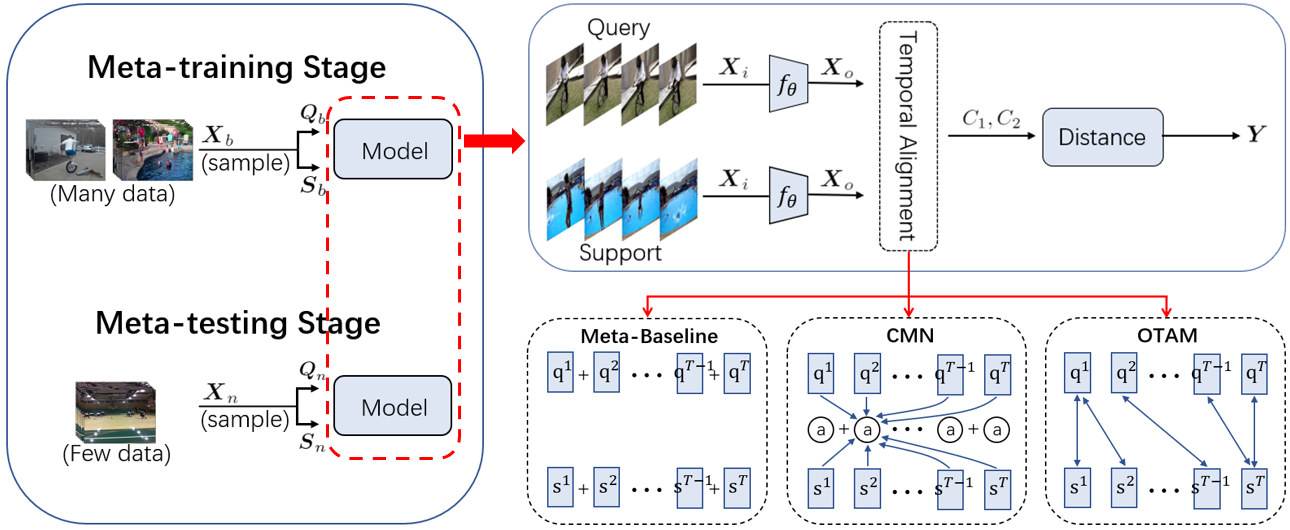}
\vspace{-3mm}
\caption{{\bf Meta-learning few-shot video classification methods.} Most existing methods employ a metric learning approach. Different methods have different temporal alignment designs. We consider three different designs in our study and only show a pair of samples from ${\bf Q}_b$ and ${\bf S}_b$ for simplicity.} 

\label{fig:meta_learning}
\vspace{-5mm}
\end{figure}

We first revisit several meta-learning based few-shot video classification methods. In meta-learning paradigm, we have abundant base class labeled data ${\bf X}_b$ and a small amount of novel class labeled data ${\bf X}_n$. The goal is to train a network that can generalize well to novel classes with limited labeled samples. As shown in Figure \ref{fig:meta_learning}, meta-learning methods consist of a meta-training and a meta-testing stage. In the meta-training stage, a collection of episodes will be randomly sampled, each of which consists of a labeled support set ${\bf S}_b$ and an unlabeled query set ${\bf Q}_b$ from ${\bf X}_b$. Specifically, in a $n$-way, $k$-shot problem, the support set consists of $n \times k$ labeled samples ($n$ samples per class). The objective is to minimize prediction loss of the samples in ${\bf Q}_b$. In the meta-testing stage, ${\bf X}_b$ is replaced with ${\bf X}_n$, and the prediction is based on unseen classes during meta-training.

In the meta-learning paradigm, metric based methods are commonly used in few-shot video classification. As shown in Figure \ref{fig:meta_learning}, a fixed number of frames $\mathbf{X}_i \in \mathbb{R}^{C_{in} \times T \times H \times W}$ are sampled sparsely and a 2D feature extractor $f_\theta$ is used to extract features $\mathbf{X}_o \in \mathbb{R}^{C \times T}$. Here, we denote the frame resolution by $H \times W$, the dimension by $C$, the number of frames by $T$, and the $k^{th}$ frame of $\mathbf{X}_o$ by $\mathrm x^k$. After temporal alignment, features of different samples $C_1, C_2$ are directly used to calculate the distance for classification. 
Different few-shot video classification methods differ in their temporal alignment methods. In our study, we consider three popular temporal alignment methods: implicit temporal alignment (CMN \cite{CMN}), explicit temporal alignment (OTAM \cite{OTAM}), and no temporal alignment (Meta-Baseline \cite{proto}). 
CMN introduces a multi-saliency embedding module to detect different salient parts of a video, which maps the original sequence to a multi-saliency descriptor. This descriptor is then used as the video level feature, which implicitly aligns two video sequences. 
OTAM explicitly matches two video sequences using a variant of the Dynamic Time Warping algorithm. The features of two aligned videos are used to measure the similarity. 
In addition, we also consider a simple meta-baseline without temporal alignment, which simply averages over the temporal dimension of frame-level features. 
Note that none of these methods contain a classifier.

\subsection{Our Method} \label{our_method}

\begin{figure}[t]

\centering
\includegraphics[scale=0.4]{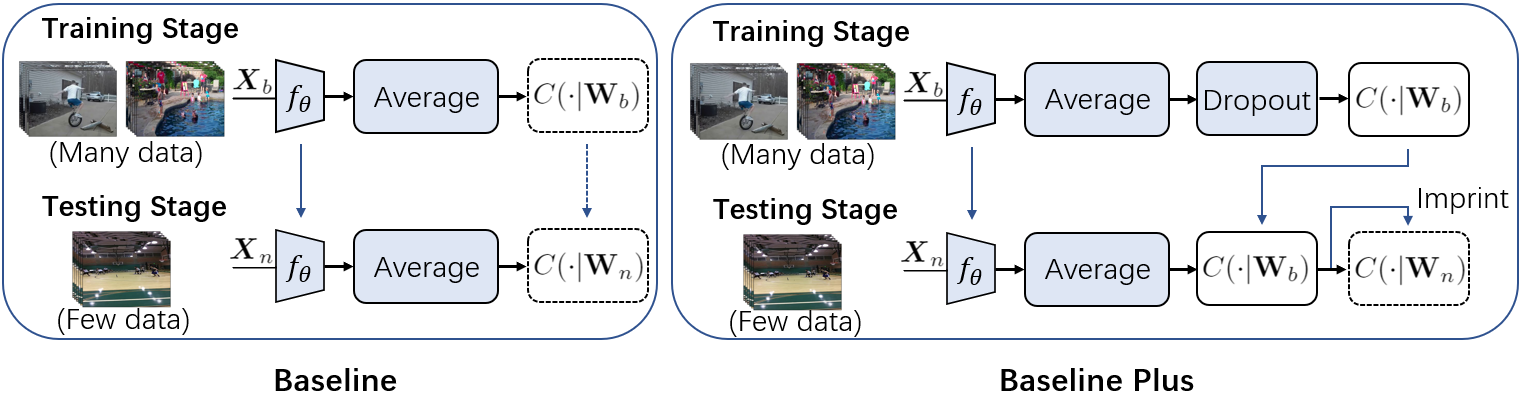}
\vspace{-4mm}
\caption{{\bf Baseline and Baseline Plus few-shot video classification methods.} Both Baseline and Baseline Plus train a feature extractor $f_\theta$ with a \emph{linear} classifier $C(.|{\bf W}_b)$ in the training stage. Baseline Plus adopts dropout to improve generalization performance and weight imprinting to help train a new classifier with limited novel data.}
\label{fig:baseline}
\vspace{-4mm}
\end{figure}

{\bf Baseline.} As shown in Figure~\ref{fig:baseline}, from a {\em pre-training and fine-tuning} perspective on few-shot recognition, we present a classifier-based baseline method with focus on learning effective representations. In the training stage, we use TSN with the same feature extractor $f_\theta$ as the meta-learning methods. For temporal aggregation, we simply average over the temporal dimension of $f_\theta(x)$ just like Meta-Baseline. We train a linear classifier $C(.|{\bf W}_b)$ by minimizing a standard cross-entropy classification loss using all the base class data ${\bf X}_b$. In the testing stage, we fix $f_\theta$ and train a new linear classifier $C(.|{\bf W}_n)$ using only labeled data in the support set (e.g., five samples are used to train the new classifier for a 5-way 1-shot task).

\label{baseline plus}
\noindent {\bf Baseline Plus.} Inspired by several improved baseline methods \cite{baseline,closer,metabaseline,imprint}, we propose our modified baseline for few-shot video classification, termed as Baseline Plus.

\noindent {\bf Training.} In the training stage of Baseline Plus, we hope to train a better feature extractor $f_\theta$ with stronger generalization ability than Baseline. So, we add a dropout layer before the linear classifier without any other modification. Dropout is an effective way to improve the generalization ability of the model, which is of great importance for few-shot learning. However, dropout will bring a negative effect on metric based methods in the meta-learning paradigm. Intuitively, dropout between the embedding and a linear classifier forces the network to update a sub-network during training, and the average of a series of sub-networks during testing (set dropout to 0) can achieve better generalization performance. However, for metric based methods, setting dropout to 0 is not equivalent to the average result of all sub-networks due to the normalization.

\noindent {\bf Testing.} In the testing stage of Baseline Plus, we fix $f_\theta$ and hope to improve the classifier with only limited novel data. Inspired by weight imprinting \cite{baseline,imprint} and cosine similarity classifier \cite{closer}, we use a similar technique to train a {\em new linear classifier} with {\em better initialization} compared to random initialization. As shown in Figure \ref{fig:baseline}, we retain the classifier $C(.|{\mathbf W}_b)$ and append a new linear classifier $C(.|{\mathbf W}_n)$ that takes the logits as input, which are better clustered than features \cite{logit,baseline}. Next, we discuss the details of 
imprinting. For a sample $({\mathbf x},y)$, denote the class number of training and testing by $C_{train}$ and $C_{test}$ respectively; logits by $ \mathbf {z( x};y) \in \mathbb{R}^{C_{train}}$; the weights and biases of $C(.|{\mathbf W}_n)$ by $\mathbf{w} \in \mathbb{R}^{C_{train} \times C_{test}}$ and $\mathbf{b} \in \mathbb{R}^{C_{test}}$ respectively; and the $k^{th}$ column of $\mathbf{w}$ by $\mathbf{w}_k$. We imprint $\mathbf{w}_k$ with $ \mathbf {z( x}_k;k) / \lVert \mathbf {z( x}_k;k) \rVert$ and $\mathbf{b}$ with $0$ where $\mathbf {x}_k$ denotes the $k^{th}$ sample in the support set for 1-shot tasks. For multiple-shot tasks, we average the logits $ \mathbf {z( x};y)$ of each class. Intuitively, this imprinted weights $\mathbf{w}$ of $C(.|{\mathbf W}_n)$ can be seen as the templates of support samples. When a query sample passes the classifier, a higher score means more similarity to the corresponding template, which is better than random initialization.

\noindent {\bf Discussion.} The combination of the idea of metric learning and classifiers can effectively improve the generalization ability in few-shot image classification. However, few-shot video classification is more challenging so the ability to learn feature representations becomes more crucial. The cosine similarity classifier does not necessarily improve the ability to learn feature representations, and it also conflicts with dropout as well as the metric based methods. We want to introduce the idea of metric learning while preserving the linear classifier and dropout, so we improve the generalization ability of the linear classifier by weight imprinting.

Our Baseline Plus is motivated by previous attempts in few-shot image classification with minor differences. In particular, Chen \etal~\cite{closer} introduced a cosine similarity classifier to reduce intra-class variations among features. Qi \etal~\cite{imprint} normalized both embeddings and columns of the weight matrix in the last layer to measure cosine similarity. Dhillon \etal~\cite{baseline} adopted transductive fine-tuning, which changes the embedding dramatically. In our case, we use a linear classifier without measuring cosine similarity and fix the embedding during testing, hoping to provide a good initialization for the classifier.

\subsection{Discussion on ImageNet Pre-training}\label{using}

\begin{figure}[t]

\centering
\includegraphics[scale=0.35]{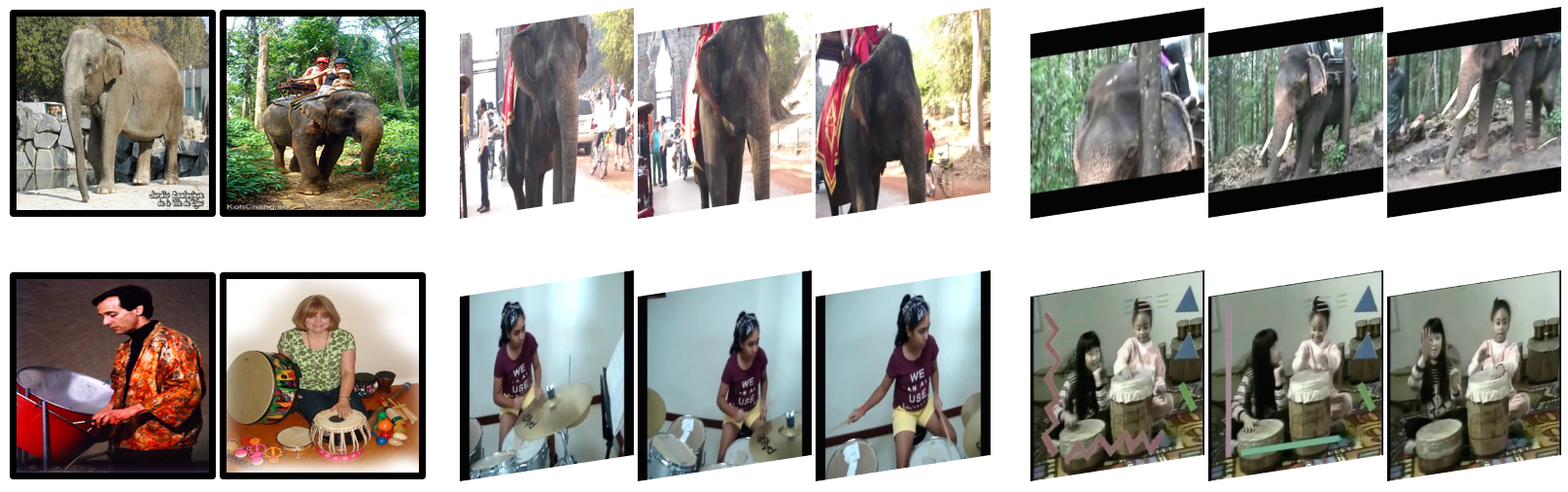}
\vspace{-4mm}
\caption{{\bf Examples of similar classes between the Kinetics and ImageNet.} The first two images in the first and second lines are from "385.Indian elephant" and "541.drum" of the ImageNet dataset, respectively. The two videos (3 frames each) in the first and second lines are from "094.riding\_elephant" and "090.playing\_drums" of the Kinetics dataset, respectively. These visualizations show that there is a strong similarity between the novel classes of the Kinetics and the classes of the ImageNet, even if their class names are different.}
\label{fig:similar}
\vspace{-4mm}
\end{figure}

Almost all existing few-shot video classification methods \cite{CMN,TARN,OTAM,TRX,implicit,PAL,CMOT} use pre-trained weights (ImageNet or Sports-1M). 
However, this pre-training step might transfer the semantic knowledge learned from the ImageNet or Sports-1M to downstream few-shot learning. This transfer process might be problematic as the novel classes might very similar to the pre-training sets, thus violating the assumption that novel classes cannot be seen during the training phase. As shown in Figure \ref{fig:similar}, some novel classes of the Kinetics are very close to some classes of the ImageNet. However, the key idea of few-shot learning is that the network should learn to generalize to \textbf{\emph{novel classes}}  (unseen during training) rather than \textbf{\emph{new samples}}. In contrast, the pre-trained network may have seen hundreds of samples that belong to novel classes through pre-trained weights, which may not reveal the real generalization performance and violates the principle of few-shot learning.

\section{Experimental Results}

\subsection{Experimental Setup}

{\bf Datasets.} Few-shot versions of the Kinetics \cite{kinetics} and the Something-Something V2 \cite{something} datasets are commonly used to evaluate few-shot video classification methods.
We first use the same splits and samples as the previous work. For the Kinetics dataset\footnote {\scriptsize  \url{https://github.com/ffmpbgrnn/CMN/tree/master/kinetics-100}}, 64/12/24 non-overlapping classes and 6,400/1,200/2,400 videos are used for training/validation/testing respectively. For the Something-Something V2 dataset\footnote {\scriptsize  \url{https://drive.google.com/drive/u/1/folders/1eyQmM2ZPXYOH_tuvseFP7yHg7tnuixqw}}, 64/12/24 non-overlapping classes and 67,013/1,926/2,857 videos are used for training/validation/testing respectively. Then, in section \ref{newbenchmark}, we propose a new benchmark to report the results without pre-training.

\noindent {\bf Implementation Details.} In each episode, we randomly sample $n$ classes with each class containing $k$ labeled instances as our support set and 1 instance as our query set. In the testing stage, we report the mean accuracy by randomly sampling 10,000 episodes in the experiments. For Baseline and Baseline Plus, we only use the support set ($n \times k$ labeled instances) to train a new classifier for 100 iterations. For meta-learning methods, details have been described in Section \ref{meta_methods}. 

We follow the video prepossessing procedure introduced in TSN \cite{TSN}. During training, we apply standard data augmentation including random crop, flip, and jitter. We crop a $224 \times 224$ region and sample $T=8$ frames per video in all stages. We use the Adam optimizer with an initial learning rate $10^{-3}$ for all methods without using pre-trained weights and a smaller initial learning rate ($10^{-4}$ for classifier-based methods and $10^{-5}$ for meta-learning methods) when using pre-trained weights. We use the validation set to select the training episodes with the best accuracy.

Some implementation details are slightly different. For CMN, we only use the multi-saliency embedding module and 8 frames as input to get the flattened feature. We initialize the hidden variable diagonally with a small constant, which is approximately equivalent to using average pooling as initialization. For OTAM, we directly use the result of DTW \cite{DTW} as the index instead of backpropagating the gradient through SoftDTW.

\subsection{Results on the Existing Benchmarks}\label{existing}

In this section, we conduct experiments on the existing benchmarks, i.e., we train a ResNet-50 network pre-trained on the ImageNet dataset to encode frames. By default, we conduct the 5-way few-shot classification.

From Table \ref{kinetics_table} and \ref{something_table}, we observe that CMN and OTAM outperform Meta-Baseline in all experiments. However, the performance gap between different temporal alignment methods is reduced in this fair comparison, in particular for 5-shot recognition. Our re-implementation of existing work improves the performance of some of the methods mainly because we used a smaller learning rate and adopt temporal jittering for data augmentation. We also improve the results of CMN by initializing the hidden variable diagonally with a small constant. On the other hand, our re-implementation of OTAM is a little lower than the reported results, which can be attributed to the modifications of some implementation details (e.g., optimizer) to ensure a fair comparison. 

\begin{figure}[t]

\centering
\includegraphics[scale=0.35]{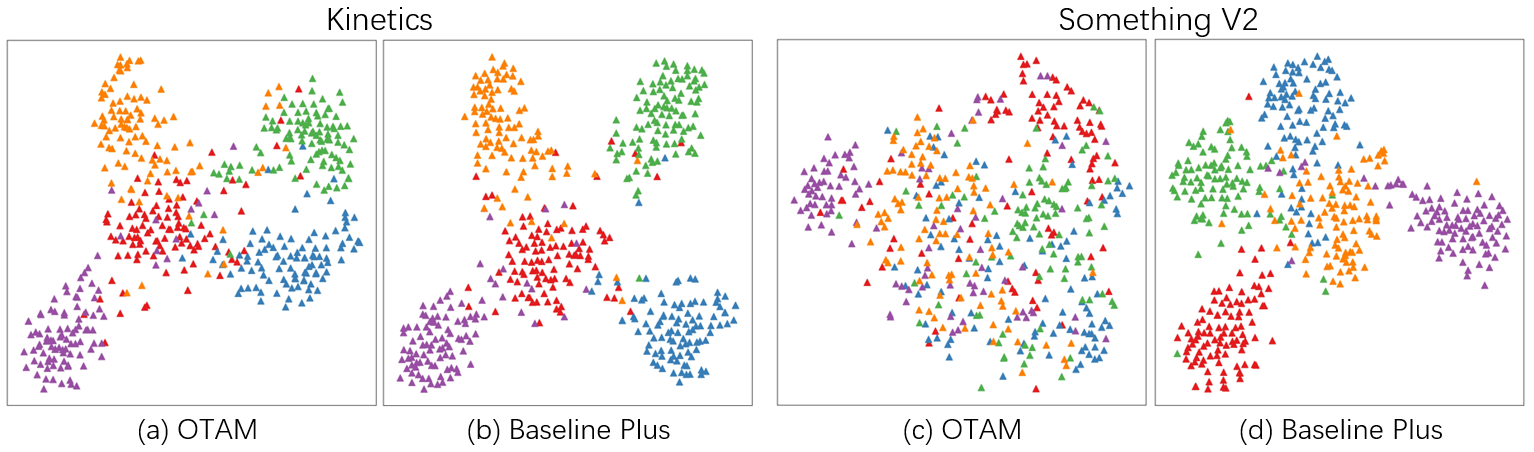}
\vspace{-4mm}
\caption{{\bf t-SNE visualization of distribution on the \emph{training set}.} The results on the Kinetics dataset and the Something v2 dataset are shown in figure (a)(b) and figure (c)(d), respectively. Different colors represent different classes. `$\blacktriangle$' in figure (a)(c) and (b)(d) represents features extracted by the feature extractor $f_\theta$ of OTAM and Baseline Plus, respectively.}
\label{fig:tsne}
\vspace{-4mm}
\end{figure}

\begin{table}[]
\begin{center}
\resizebox{0.75\textwidth}{!}{
\begin{tabular}{lcccc}
\hline
              & \multicolumn{2}{c}{1-shot} & \multicolumn{2}{c}{5-shot} \\
Method        & Reported   & Ours          & Reported   & Ours          \\ \hline
Meta-Baseline \cite{proto}* & 64.5       & 64.03 $\pm$ 0.41  & 77.9       & 80.43 $\pm$ 0.35  \\
CMN \cite{CMN}          & 60.5       & 65.90 $\pm$ 0.42  & 78.9       & 82.72 $\pm$ 0.34  \\
OTAM \cite{OTAM}         & 73.0       & 71.45 $\pm$ 0.40 & 85.8       & 82.10 $\pm$ 0.34 \\ \hline
Baseline (ours)      & -          & 69.48 $\pm$ 0.39  & -          & 84.41 $\pm$ 0.33  \\
Baseline Plus (ours) & -          & \textbf{74.63 $\pm$ 0.37}  & -  & \textbf{86.62 $\pm$ 0.26}  \\ \hline
\end{tabular}
}
\end{center}
\vspace{-3mm}
\caption{{\bf Few-shot video classification results on the Kinetics dataset.} We report the mean of 10,000 randomly generated test episodes as well as the 95\% confidence intervals. We use a ResNet-50 backbone and ImageNet pre-trained weights for all methods. *: Results reported in \cite{OTAM}.}
\label{kinetics_table}
\vspace{-3mm}
\end{table}

\begin{table}[t]
\begin{center}
\resizebox{0.75\textwidth}{!}{
\begin{tabular}{lcccc}
\hline
              & \multicolumn{2}{c}{1-shot} & \multicolumn{2}{c}{5-shot} \\
Method        & Reported   & Ours          & Reported   & Ours          \\ \hline
Meta-Baseline \cite{proto} & 33.6       & 37.31 $\pm$ 0.41  & 43.0       & 48.28 $\pm$ 0.44  \\
CMN \cite{CMN}           & 34.4       & 40.62 $\pm$ 0.42  & 43.8       & 51.90 $\pm$ 0.44  \\
OTAM \cite{OTAM}          & 42.8       & 41.55 $\pm$ 0.42 & 52.3       & 51.33 $\pm$ 0.43 \\ \hline
Baseline (ours)     & -          & 40.78 $\pm$ 0.41 & -          & 59.21 $\pm$ 0.44 \\
Baseline Plus (ours) & -         & \textbf{46.04 $\pm$ 0.42}  & -    & \textbf{61.10 $\pm$ 0.39}  \\ \hline
\end{tabular}
}
\end{center}
\vspace{-3mm}
\caption{{\bf Few-shot video classification results on the Something V2 dataset.} All experimental settings are the same as the Kinetics dataset.}
\label{something_table}
\vspace{-3mm}
\end{table}

\begin{table}[]
\begin{center}
\resizebox{0.75\textwidth}{!}{
\begin{tabular}{lcccc}
\hline
              & \multicolumn{2}{c}{Kinetcis}      & \multicolumn{2}{c}{Something V2}  \\
Method        & 1-shot          & 5-shot          & 1-shot          & 5-shot          \\ \hline
Meta-Baseline & 42.46 $\pm$ 0.42 & 49.78 $\pm$ 0.43 & 20.85 $\pm$ 0.33 & 21.87 $\pm$ 0.35 \\
CMN           & 40.37 $\pm$ 0.42 & 50.27 $\pm$ 0.42 & 21.10 $\pm$ 0.34 & 23.26 $\pm$ 0.36 \\
OTAM          & 44.37 $\pm$ 0.43 & 50.57 $\pm$ 0.39 & 22.75 $\pm$ 0.35 & 23.50 $\pm$ 0.32 \\ \hline
Baseline (ours)     & 44.67 $\pm$ 0.32 & 55.53 $\pm$ 0.35 & 27.06 $\pm$ 0.35 & 36.73 $\pm$ 0.42 \\
Baseline Plus (ours) & \textbf{46.24 $\pm$ 0.38} & \textbf{56.92 $\pm$ 0.37} & \textbf{36.06 $\pm$ 0.39} & \textbf{48.36 $\pm$ 0.40} \\ \hline
\end{tabular}
}
\end{center}
\vspace{-3mm}
\caption{{\bf Few-shot video classification results on both the Kinetics and Something V2 datasets without using ImageNet pre-trained weights.} Classifier-based methods surpass meta-learning methods.}
\label{nopretrain_table}
\vspace{-3mm}
\end{table}

\begin{table}[]
\vspace{-3mm}
\begin{center}
\resizebox{0.5\textwidth}{!}{
\begin{tabular}{lcc}
\hline
Method               & 1-shot       & 5-shot       \\ \hline
Meta-Baseline        & 59.95 $\pm$ 0.42 & 71.43 $\pm$ 0.40 \\
CMN                  & 61.20 $\pm$ 0.42 & 73.46 $\pm$ 0.39 \\
OTAM                 & 64.56 $\pm$ 0.41 & 74.67 $\pm$ 0.37 \\ \hline
Baseline (ours)      & 62.02 $\pm$ 0.42 & 75.23 $\pm$ 0.38 \\
Baseline Plus (ours) & \textbf{65.76 $\pm$ 0.42} & \textbf{76.94 $\pm$ 0.36} \\ \hline
\end{tabular}
}
\end{center}
\caption{{\bf Few-shot video classification results on the Kinetics dataset using MoCo pre-trained weights.} All methods become worse compared to their supervised pre-training counterparts.}
\label{moco_table}
\end{table}

\begin{table}[]
\begin{center}
\resizebox{0.80\textwidth}{!}{
\begin{tabular}{lcccc}
\hline
                     & \multicolumn{2}{c}{Kinetics} & \multicolumn{2}{c}{Kinetics (ImageNet)} \\ 
Method               & 1-shot        & 5-shot       & 1-shot             & 5-shot             \\ \hline
cosine classifier \cite{closer}           & 44.43 $\pm$ 0.37  & 54.13 $\pm$ 0.40 & 62.46 $\pm$ 0.40       & 81.38 $\pm$ 0.33       \\
weight imprinting \cite{imprint}    & 45.11 $\pm$ 0.37  & 55.92 $\pm$ 0.40 & 64.47 $\pm$ 0.41       & 82.04 $\pm$ 0.34       \\
Baseline (ours)      & 44.67 $\pm$ 0.32  & 55.53 $\pm$ 0.35 & 69.48 $\pm$ 0.39       & 84.41 $\pm$ 0.33       \\
Baseline Plus (ours) & \textbf{46.24 $\pm$ 0.38}  & \textbf{56.92 $\pm$ 0.37} & \textbf{74.63 $\pm$ 0.37}       & \textbf{86.62 $\pm$ 0.26}       \\ \hline
\end{tabular}
}
\end{center}
\vspace{-3mm}
\caption{{\bf Comparison of different classifier-based methods on the Kinetics dataset.} (ImageNet) means using ImageNet pre-trained weights as initialization.}
\label{baseline_table}
\vspace{-3mm}
\end{table}

\begin{table}[t]
\begin{center}
\resizebox{0.80\textwidth}{!}{
\begin{tabular}{lcccc}
\hline
              & \multicolumn{2}{c}{Kinetics} & \multicolumn{2}{c}{complete-Kinetics} \\
Method                                          & 1-shot                                     & 5-shot                                     & 1-shot                                     & 5-shot                                     \\ \hline
Meta-Baseline                                   & 42.46 $\pm$ 0.42                           & 49.78 $\pm$ 0.43                           & 42.22 $\pm$ 0.42                           & 53.64 $\pm$ 0.44                           \\
CMN                                             & 40.37 $\pm$ 0.42                           & 50.27 $\pm$ 0.42                           & 43.45 $\pm$ 0.42                           & 51.89 $\pm$ 0.43                           \\
OTAM                                            & 44.37 $\pm$ 0.43                           & 50.57 $\pm$ 0.39                           & 46.41 $\pm$ 0.42                           & 52.05 $\pm$ 0.39                           \\ \hline
Baseline (ours)                                 & 44.67 $\pm$ 0.32                           & 55.53 $\pm$ 0.35                           & 49.44 $\pm$ 0.42                           & 64.22 $\pm$ 0.44                           \\
Baseline Plus (ours)    & \textbf{46.24 $\pm$ 0.38} & \textbf{56.92 $\pm$ 0.37} & \textbf{58.34 $\pm$ 0.38} & \textbf{69.90 $\pm$ 0.35} \\ \hline
\end{tabular}
}
\end{center}
\vspace{-3mm}
\caption{{\bf Few-shot video classification results on the complete-Kinetics dataset without using ImageNet pre-trained weights.} Our classifier-based methods gain a large performance improvement compared to meta-learning methods.} 
\label{newbenchmark_table}
\vspace{-3mm}
\end{table}

The performance of the classifier-based methods is surprising. On both datasets of the Kinetics and the Something V2, even Baseline achieves competitive performance to other meta-learning methods. Baseline Plus further improves Baseline and significantly outperforms state-of-the-art meta-learning methods. Since both Baseline and Baseline Plus simply average extracted frames, the strong performance could only be attributed to their ability to learn feature representation, which has been neglected by previous work. As shown in Figure \ref{fig:tsne}, OTAM has a weaker feature representation compared with Baseline Plus, which should be attributed to the difference in the training process of meta-learning methods and classifier-based methods. Although the meta-learning paradigm is designed for learning to learn, the ability of effectively learning low-level features (e.g., how to encode a sample rather than complete an episode) becomes more critical on a deeper backbone and a more challenging video classification task. To verify this conjecture, we conduct experiments to investigate the effects of not using pre-trained weights in the next section. 

\subsection{Study on Different Pre-training Strategies}\label{pretraining}

In this section, we first train all the methods from scratch without using ImageNet pre-trained weights, which is more reasonable as discussed in Section \ref{using}. We use the same Kinetics and Something V2 datasets. The only difference is that we replace the initialization from ImageNet pre-trained weights with random initialization. The results on both the Kinetics and the Something-Something V2 datasets are listed in Table \ref{nopretrain_table}. The performance of all methods greatly decrease. Both Baseline and Baseline Plus can still outperform all meta-learning methods under this scenario. We see that the performance of meta-learning methods is similar to random guess (20\%) on 5-way task in Something v2 dataset, which might be due to the incapacity of learning effective features.

To clarify whether the better performance of ImageNet supervised pre-training is related to overlapping information from ImageNet or simply due to the use of pre-training on a large dataset, we conduct experiments on the Kinetics dataset using self-supervised pre-training. MoCo \cite{moco} is designed for unsupervised visual representation learning and can outperform its ImageNet supervised pre-training counterpart in some downstream tasks, which suggests that we should get comparable or better results if we use it as pre-training. However, as shown in Table \ref{moco_table}, all methods are worse compared to their supervised pre-training counterpart (Table \ref{kinetics_table}). We believe this performance gap is due to the fact that the supervised pre-training introduces labels in the pre-training stage, and in the presence of overlapping information from ImageNet, the supervised pre-trained models learn more knowledge related to novel classes than the self-supervised approach. The above results show that the performance improvement using pre-trained weights is partly due to the introduction of 
novel classes, which violates the principle of few-shot learning. In addition, we notice that even unsupervised MOCO pre-training can greatly improve the performance over the training from scratch, which further confirms our assumption that representation learning is still the main challenge for few-shot video classification, and we should pay more attention on designing effective pre-training strategies in video domain for better few-shot video classification.

\subsection{Comparison with Other Classifier-based Methods}
Since previous classifier-based methods are all designed for images, we compare Baseline Plus to two common methods in few-shot image classification (cosine classifier \cite{closer} and weight imprinting \cite{imprint}), which have been discussed in Section \ref{our_method}. Note that they both use cosine classifiers or their variants. From Table \ref{baseline_table}, we can conclude that the cosine classifier is not necessarily better than the linear classifier, especially when we use pre-trained weights. Our Baseline Plus introduces the idea of metric learning while preserving the linear classifier and dropout, which makes it surpass other classifier-based methods.

\subsection{A New Benchmark}\label{newbenchmark}

Furthermore, there is no reason to limit the size of the training set since we should be provided with abundant labeled data during training in the few-shot learning setting. Both our experimental results in section \ref{pretraining} and previous work \cite{imagenet} reveal that a small training set can lead to overfitting without using pre-trained weights. Therefore, we propose a new benchmark with sufficient training data, termed as {\em complete-Kinetics}. We use the same training/validation/testing splits and the same validation/testing samples as the Kinetics dataset. The only difference is that we use all the training samples from the original Kinetics dataset, that is, 64/12/24 non-overlapping classes and 49,325/1,200/2,400 videos are used for training/validation/testing respectively. 

The results on this new benchmark are shown in Table \ref{newbenchmark_table}. 
Compared with the results on the Kinetics dataset, the performance of Baseline and Baseline Plus improves by a large margin without using pre-trained weights, which confirms the significance of sufficient training data. In contrast, the improvement of meta-learning methods is relatively limited. Thus, we believe that how to improve the ability of meta-learning methods to effectively learn feature representation is an important future direction, especially on challenging tasks.  

\vspace{-5mm}
\section{Conclusion}

In this paper, we have first compared the existing meta-learning methods of few-shot video classification and present a new classifier-based model Baseline Plus. Surprisingly, this simple baseline outperforms the state-of-the-art meta-learning methods on the existing benchmarks. Furthermore, we rethink the reasonableness of the existing benchmarks and compare different pre-training strategies. Through a fair comparison, our results reveal that the main bottleneck of few-shot video classification is the ability of learning feature representation, and the existing datasets lack enough base data for deep model training. Thus we establish a new benchmark with more based data, termed as {\em complete-Kinetics}. We hope that this new baseline and benchmark can provide some insights for the future research of few-shot video classification.

\paragraph{\bf Acknowledgements.} Limin Wang is the corresponding author. This work is supported by the National Natural Science Foundation of China (No. 62076119), Program for Innovative Talents and Entrepreneur in Jiangsu Province, and Collaborative Innovation Center of Novel Software Technology and Industrialization.

\bibliography{egbib}
\end{document}